\documentclass[11pt,leqno]{amsart}
\usepackage{t1enc}
\usepackage[latin1]{inputenc}
\usepackage[english]{babel}

\usepackage{booktabs}
\usepackage{multirow}
\usepackage{amssymb, mathtools, color, ytableau, bbm}
\ytableausetup{smalltableaux}
\usepackage[colorinlistoftodos]{todonotes}
 \usepackage{hyperref}
 \hypersetup{
     colorlinks=true,
     linkcolor=blue,
     filecolor=magenta,
     citecolor = blue,
     urlcolor=blue,
 }

\makeatletter
\newcommand\notsotiny{\@setfontsize\notsotiny\@vipt\@viipt}
\makeatother

\theoremstyle{definition}
\newtheorem {theorem}{Theorem}[section]
\newtheorem {lemma}[theorem]{Lemma}

\newtheorem{definition}[theorem]{Definition}
\newtheorem{remark}[theorem]{Remark}
\newtheorem{example}[theorem]{Example}

\newenvironment{red}{\relax\color{red}}{\relax}
\newenvironment{blue}{\relax\color{blue}}{\hspace*{.5ex}\relax}

\newcommand{\ber}{\begin{red}}
\newcommand{\er}{\end{red}}
\newcommand{\beb}{\begin{blue}}
\newcommand{\eb}{\end{blue}}

\newcommand{\seteq}{\coloneqq}

\newcommand{\GL}{\operatorname{GL}}

\numberwithin{equation}{section}

\begin{document}

\title[Interpretable Machine Learning Kronecker Coefficients]{Interpretable Machine Learning \\ for Kronecker Coefficients}

\author[G.\ Butbaia, K.-H.\ Lee and F.\ Ruehle]{Giorgi Butbaia, Kyu-Hwan Lee and Fabian Ruehle}

\date{\today}
%\keywords{Kronecker coefficients, principal component analysis, kernel methods}

\begin{abstract}
We analyze the saliency of neural networks and employ interpretable machine learning models to predict whether the Kronecker coefficients of the symmetric group are zero or not. Our models use triples of partitions as input features, as well as $b$-loadings derived from the principal component of an embedding that captures the differences between partitions. Across all approaches, we achieve an accuracy of approximately 83\% and derive explicit formulas for a decision function in terms of $b$-loadings. Additionally, we develop transformer-based models for prediction, achieving the highest reported accuracy of over 99\%.
\end{abstract}

\maketitle

\section{Introduction}\label{sec1}

This article is a sequel to \cite{L1,L2}, where the second author studied Kronecker coefficients using data-driven approaches, including machine learning (ML) models. %(See also \cite{DouglasLee2025} in this volume for an expository discussion.) 
In particular, \cite{L1} applies convolutional neural networks (CNNs) and LightGBM to datasets of Kronecker coefficients, demonstrating that the trained classifiers achieve high accuracy ($\approx$ 98\%) in predicting whether a Kronecker coefficient vanishes. These results suggest that further data-driven analysis may uncover new structural insights into Kronecker coefficients. Indeed, a subsequent paper \cite{L2} employs principal component analysis (PCA) and kernel methods to formulate a sufficient condition for the nonvanishing of Kronecker coefficients, providing an effective tool for their study. However, the underlying reasons for the high accuracy of ML models remain largely unexplained, as CNNs and LightGBM are not easily interpretable. 

In this article, we employ various interpretable approaches to whitebox the ML results on the binary classification task of predicting which Kronecker coefficients of the symmetric group $\mathfrak S_n$ are zero. First, we perform a gradient saliency analysis similar to \cite{DVB+} on a classifier network. Second, we use Kolmogorov--Arnold Networks (KANs), developed by the third author in collaboration \cite{liu2024kan}, both to examine the learned function and as a symbolic regressor. Third, we train a very small neural network and analyze the parameters of the learned function. Finally, we apply PySR~\cite{cranmer2023interpretable} for symbolic regression on the datasets.

Our analysis and models use triples of partitions of $n$ as input, as well as the \textit{$b$-loading} of partitions, introduced in \cite{L2} as a result of PCA applied to an embedding that captures differences in partitions. Both the saliency analysis and KANs reveal that when a partition is represented as an $n$-dimensional vector, the first few and last few entries are more significant than the middle ones. When using $b$-loadings, all three interpretable models yield comparable results, achieving approximately 83\% accuracy in each case (the theoretical upper bound on the precision obtainable from $b$-loadings is around 85\%). Since these models are interpretable, we can extract explicit formulas for their decision functions.

Additionally, beyond the main theme of this article, we include results from transformer models, which achieve the highest reported accuracy of over 99\%. Understanding the interpretability of these transformer models and its implications for the computational complexity or approximation of the underlying decision problem would be an interesting direction for future research.

\subsection*{Acknowledgments}
Lee is partially supported by a grant from the Simons Foundation (\#712100). The work of Ruehle is supported by the NSF grants PHY-2210333 and PHY-2019786 (The NSF AI Institute for Artificial Intelligence and Fundamental Interactions), as well as startup funding from Northeastern University.  We are grateful for support from CMSA for the Mathematics and Machine Learning Program during the Fall 2024 semester, where this project was initiated.

\section{Preliminaries}\label{s:Ltraining}
\subsection{Kronecker coefficients}
  
Let $\mathfrak S_n$ be the symmetric group of degree $n$. The irreducible representations $S_\lambda$ of $\mathfrak S_n$ over $\mathbb C$ are parametrized by partitions $\lambda$ of $n$, written as $\lambda \vdash n$. The tensor product of two irreducible representations $S_\lambda$ and $S_\mu$ ($\lambda, \mu \vdash n$) is decomposed into a sum of irreducible representations:
\[  S_\lambda \otimes S_\mu = \bigoplus_{\nu\vdash n} g_{\lambda, \mu}^\nu S_\nu  \quad (g_{\lambda, \mu}^\nu \in \mathbb Z_{\ge 0}) . \]
The decomposition multiplicities $g_{\lambda, \mu}^\nu$ are called the {\em Kronecker coefficients}.

In stark contrast to the analogous Littlewood--Richardson coefficients for $\GL_N(\mathbb C)$, no combinatorial description of $g_{\lambda, \mu}^\nu$ has been found since Murnaghan \cite{Mur} first posed the question in 1938. This remains one of the central open problems in combinatorial representation theory, with only partial results available (e.g., \cite{Bla}). It has also been shown that determining  whether a given Kronecker coefficient is non-zero is NP-hard \cite{IMW}. 

\medskip

For later discussion, we note that there are symmetries among $g_{\lambda, \mu}^\nu$.
\begin{lemma} \cite[p.61]{FH} \label{lem-perm}
Let $\lambda, \mu, \nu \vdash n$. Then the Kronecker coefficients $g_{\lambda, \mu}^\nu$ are invariant under the permutations of $\lambda, \mu, \nu$. That is, we have
\[ g_{\lambda,  \mu}^\nu=g_{\mu, \lambda}^\nu=g_{\lambda, \nu}^\mu=g_{\nu, \lambda}^\mu=g_{\mu, \nu}^\lambda=g_{\nu, \mu}^\lambda. \]
\end{lemma}

\subsection{Different matrix and $b$-loading}

For $n \in \mathbb Z_{>0}$, let $\mathcal P(n)$ be the set of partitions of $n$.
We identify each element $\lambda$ of $\mathcal P(n)$ with a sequence of length $n$ by appending as many $0$-entries as needed.   We also consider $\mathcal P(n)$ as an ordered set by the lexicographic order. 

\begin{example} When $n=6$, we have
\begin{align*} \mathcal P(6) = \{ (6,0,0,0,0,0), (5, 1,0,0,0,0), (4, 2,0,0,0,0), (4, 1, 1,0,0,0),\\ (3,3,0,0,0,0), (3, 2, 1,0,0,0), (3, 1, 1, 1,0,0),(2,2,2,0,0,0),\\ (2, 2, 1, 1,0,0), (2, 1, 1, 1,1,0), (1,1, 1, 1, 1, 1) \} . \phantom{LLLLLLLLa} \end{align*}
\end{example}

The size of the set $\mathcal P(n)$ will be denoted by $p(n)$. Define a $p(n) \times p(n)$ symmetric matrix $\mathsf Z_n=[z_{\lambda, \mu}]_{\lambda, \mu \in \mathcal P(n)}$ by
\[ z_{\lambda, \mu}= \lVert \lambda-\mu \rVert_1 \seteq \sum_{i=1}^n | \lambda_i - \mu_i| \]
for $\lambda=(\lambda_1, \lambda_2, \dots , \lambda_n)$ and $\mu=(\mu_1,\mu_2, \dots , \mu_n) \in \mathcal P(n)$.
The matrix $\mathsf Z_n$ will be called the {\em difference} matrix of $\mathcal P(n)$.

\begin{example}
When $n=6$, we obtain 
\[{\scriptsize \mathsf Z_6 = \left [ \begin{array}{ccccccccccc}  0  &2  &4  &4  &6  &6  &6  &8  &8  &8 &10 \\ 2  &0  &2  &2  &4  &4  &4  &6  &6  &6  &8 \\  4  &2  &0  &2  &2  &2  &4  &4  &4  &6  &8\\4  &2  &2  &0  &4  &2  &2  &4  &4  &4  &6\\6  &4  &2  &4  &0  &2  &4  &4  &4  &6  &8\\6  &4  &2  &2  &2  &0  &2  &2  &2  &4  &6\\6  &4  &4  &2  &4  &2  &0  &4  &2  &2  &4\\8  &6  &4  &4  &4  &2  &4  &0  &2  &4  &6\\8  &6  &4  &4  &4  &2  &2  &2  &0  &2  &4\\8  &6  &6  &4  &6  &4  &2  &4  &2  &0  &2\\ 10  &8  &8  &6  &8  &6  &4  &6  &4  &2  &0 \end{array} \right ] }.\]  
\end{example}

Since $\mathsf Z_n$ is symmetric, all its eigenvalues are real. Moreover, the Perron--Frobenius theorem for matrices with nonnegative entries \cite[Section III.2]{Ga} guarantees that $\mathsf Z_n$ has a unique eigenvalue of largest magnitude and that the corresponding eigenvector can be chosen to have strictly positive components. Thus the following definition is well-defined.

\begin{definition}
Let $\mathbf w =(w_\lambda)_{\lambda \in \mathcal P(n)}$ be an eigenvector of the largest eigenvalue of $\mathsf Z_n$ such that $w_\lambda >0$ for all $\lambda \in \mathcal P(n)$. Denote by $w_{\mathrm{max}}$ (resp. $w_{\mathrm{min}}$) a maximum (resp. minimum) of $\{ w_\lambda \}_{\lambda \in \mathcal P(n)}$. Define \[ b_\lambda \seteq 100 \times \frac{w_\lambda - w_{\mathrm{min}}}{w_{\mathrm{max}} - w_{\mathrm{min}}} \quad \text{for } \lambda \in \mathcal P(n) .\] The value $b_\lambda$ is called the {\em $b$-loading} of partition $\lambda \in \mathcal P(n)$. 
\end{definition}

\begin{example}
When $n=6$, we have  \begin{align*} & \mathbf w \approx (0.4045,0.2961,0.2662,0.2393,\\ & \phantom{LLLLLLLLLLL} 0.3061,0.2318,0.2393,0.3061,0.2662,0.2961, 0.4045) , \end{align*}  and  the $b$-loadings are given by
\begin{align*} & (b_\lambda)_{\lambda \in \mathcal P(n)} = (100.00,37.25,19.93,4.36, \\ & \phantom{LLLLLLLLLLLLLLLLLLL} 43.01,0.00,4.36,43.01,19.93,37.25, 100.00).\end{align*}
\end{example}

\begin{remark} \label{quote-1}
The $a$-loading is also defined in \cite{L2}. However, since it is not used in this article, we do not introduce it here.   
\end{remark}

For $\mathbf t=(\lambda, \mu, \nu) \in \mathcal P(n)^3$, we will write
\[ g(\mathbf t) \seteq g_{\lambda, \mu}^\nu . \]

\begin{definition} \label{def-three}
Let $\mathbf t=(\lambda, \mu, \nu) \in \mathcal P(n)^3$. Define the {\em $b$-loading}  of $\mathbf t$, denoted by  $b(\mathbf t)$, to be the sum of the $b$-loadings of $\lambda, \mu$ and $\nu$, i.e., 
\[b(\mathbf t) \seteq b_\lambda + b_\mu + b_\nu  . \] 
\end{definition}

\begin{figure}[th]
\begin{center}
\includegraphics[scale=0.4]{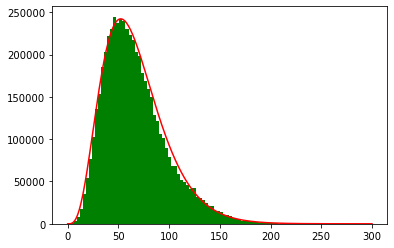}\quad \includegraphics[scale=0.4]{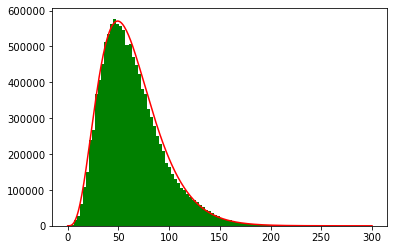}
\end{center}
\caption{\small\sf Histograms of $b$-loadings of $\mathbf t \in \mathcal P(n)^3$ for $n=15$ (left) and $16$ (right) along with curves (red) of gamma distributions.} \label{fig-7} \end{figure}

\begin{figure}[th]\begin{center}
\includegraphics[scale=0.4]{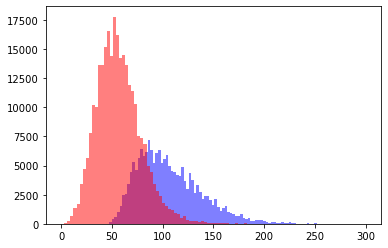}
\quad \includegraphics[scale=0.4]{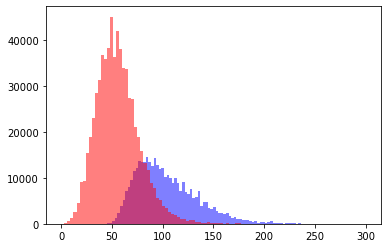}
\end{center}
\caption{\small \sf Histograms of $b$-loadings for $n=12$ (left) and $n=13$ (right). The red (resp. blue) region represents the numbers of $\mathbf t$ such that $g(\mathbf t) \neq 0$ (resp. $g(\mathbf t) =0$).} \label{fig-3} \end{figure}

The histograms of $b$-loadings $b(\mathbf t)$ clearly indicate that they follow gamma distributions (Figure \ref{fig-7}). Moreover, when examining the distributions of loadings based on whether the Kronecker coefficients $g(\mathbf t)$ are zero or nonzero (Figure \ref{fig-3}), it is evident that the minimum values of loadings can be used to define vertical lines that separate the red regions from the blue ones.

With this in mind, define
\begin{align*}   b_{\star} & \seteq \min \{ b(\mathbf t) : g(\mathbf t) = 0 , \mathbf t \in \mathcal P(n)^3 \}  . \end{align*}
Once the values of $b_\star$ is determined, this definition provides a sufficient condition for $g(\mathbf t) \neq 0$: 
\begin{equation} \label{eqn-box}   \text{For $\mathbf t \in \mathcal P(n)^3$, \quad if  $b(\mathbf t) < b_{\star}$  then $g(\mathbf t) \neq 0$ .} \end{equation}

\begin{example} \label{exa-18} \hfill
\begin{enumerate}
\item When $n=18$, we obtain $b_{\star} \approx 44.18$. Now that the $b$-loading of \[ \mathbf t=((12, 4, 2), (8, 4, 2, 2, 1, 1), (5, 4, 3, 3, 1, 1, 1))\] is readily computed to be approximately $41.07 < b_{\star}$, we immediately conclude that $g(\mathbf t) \neq 0$ by \eqref{eqn-box}.
\item When $n=20$, there are $246{,}491{,}883$ triples $\mathbf t \in \mathcal P(20)$. Among them, $78{,}382{,}890$ triples satisfy $b(\mathbf t) < b_\star \approx 43.74$ so that $g(\mathbf t) \neq 0$. The percentage of these triples is about 31.8\%.  
\end{enumerate}
\end{example}

\subsection{The threshold $b_\star$ and additional remarks}
The value $b_\star$ determines instances of $\mathbf t$ for which  $g(\mathbf t) \neq 0$ with 100\% confidence in a significant number of cases. As mentioned in the introduction, in the following sections, we will derive formulas in terms of $b$-loadings, using interpretable ML models that predict whether $g(\mathbf t) \neq 0$ with approximately 83\% accuracy.

\begin{figure}[t]
\begin{center}
\includegraphics[width=.6\textwidth, height=.4\textwidth]{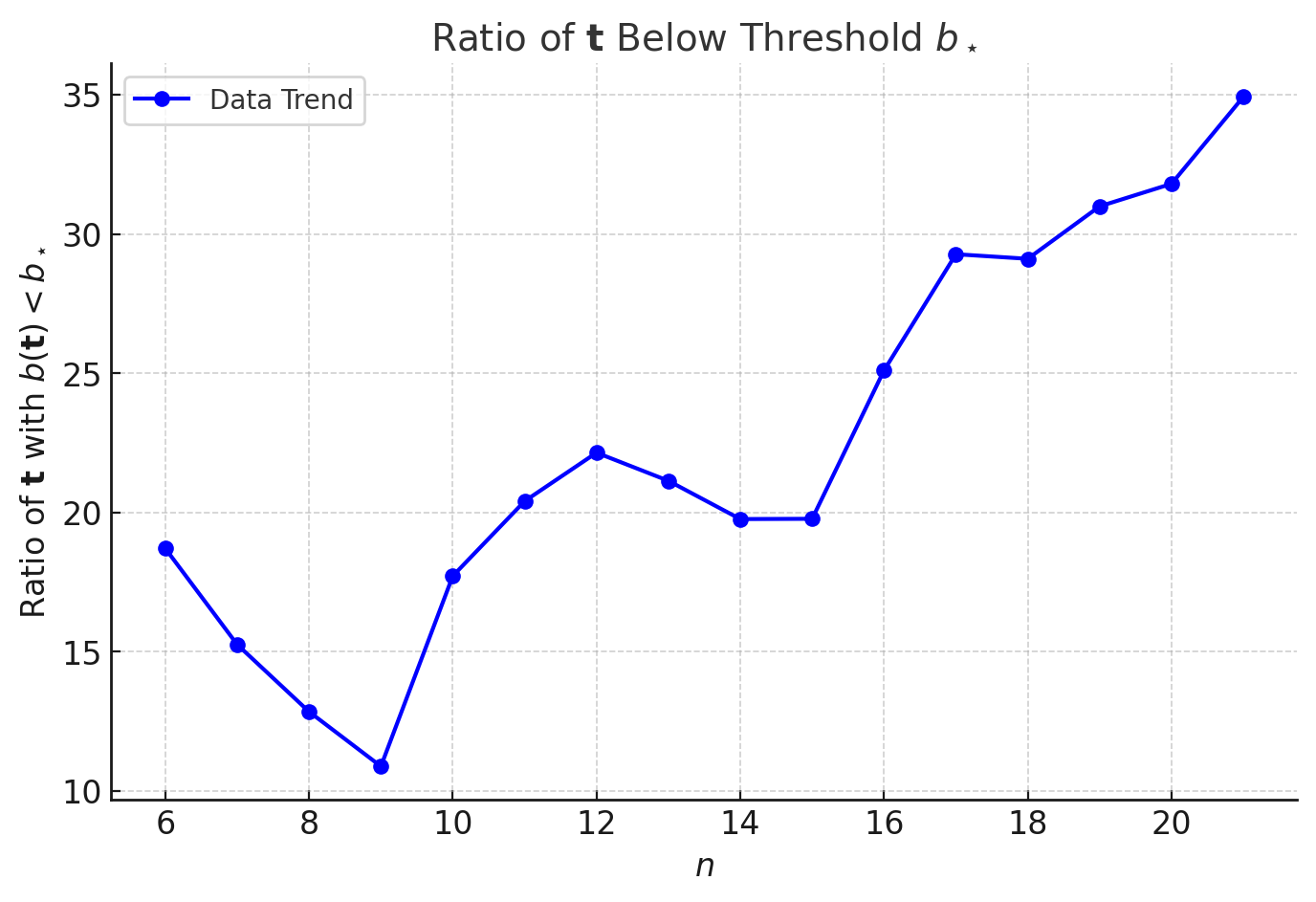}
\end{center}
\caption{\small\sf The ratio of $\mathbf t$ satisfying $b(\mathbf t) < b_\star$.} \label{fig:ratio} \end{figure}

The values of $b_\star$ are computed for $6 \le n \le 20$ in \cite{L2} and for $n=21,24,27,30,33,36$ assuming a conjecture. The ratio of $\mathbf t$ satisfying $b(\mathbf t) < b_\star$ exhibits an increasing trend for $n\ge 9$, as shown in Figure~\ref{fig:ratio}.

\begin{figure}[t]
\begin{center}
\includegraphics[width=.47\textwidth]{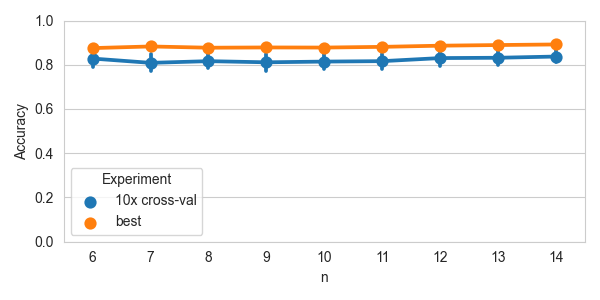}~~ \includegraphics[width=.47\textwidth]{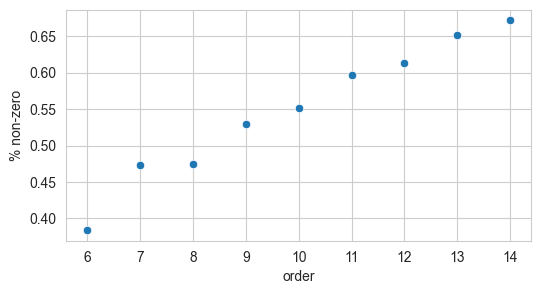}
\end{center}
\caption{\small\sf Left: Accuracies for classification based on $b$-loadings for different decision boundaries. Right: Percentage of non-zero Kronecker coefficients.} \label{fig:Accuracy} \end{figure}

The accuracy of the binary classification networks used in Section~\ref{sec:Saliency}, where the input features are the triples $\mathbf{t}=(\lambda,\mu,\nu)$ of partitions, is on a par with the accuracy obtained using $b$-loadings only. For the latter, we find an optimal decision boundary by training a binary decision tree for different values of $n$. Running the algorithm on a 10\% test set with ten-fold cross validation gives a stable performance of around 80\% across the range of $n$ studied, see the blue curve in Figure~\ref{fig:Accuracy}. The error bars indicate one-sigma confidence intervals from the cross validation. The maximum classification accuracy that can be achieved based on the $b$-loadings alone is around 85\% (see the orange line). 

In light of the NP-hardness of the decision problem, this suggests that the average-case complexity is not NP-hard, since computing the $b$-values is polynomial, and the data science methods we employ run in polynomial time. Finally, we want to point out that the accuracy of predicting zero vs. non-zero coefficients remains almost constant, albeit the percentage of non-zero coefficients varies (see the right plot of Figure~\ref{fig:Accuracy}). At $n=9$, the majority class shifts from zero to non-zero coefficients. However, the classifier accuracy is unaffected by this.

\newpage

\subsection{Data structures}
\subsubsection{$3n$-dimensional input} \label{subsub-3n}
\begin{figure}[t]
\begin{center}
\includegraphics[width=.3\textwidth]{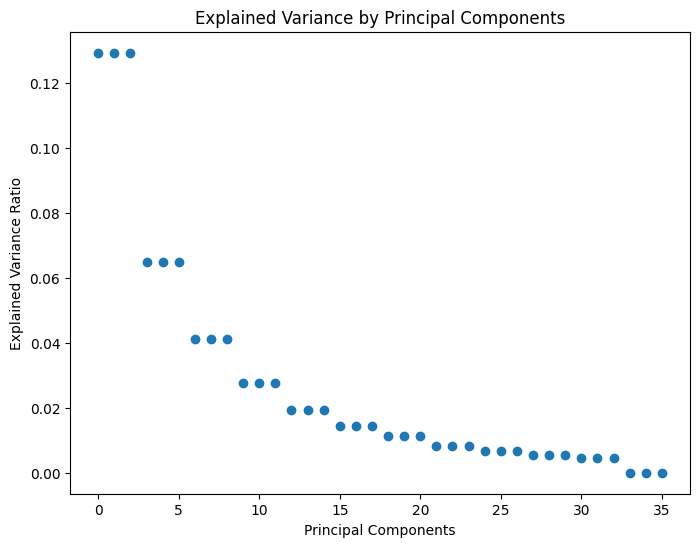}~~ \includegraphics[width=.3\textwidth]{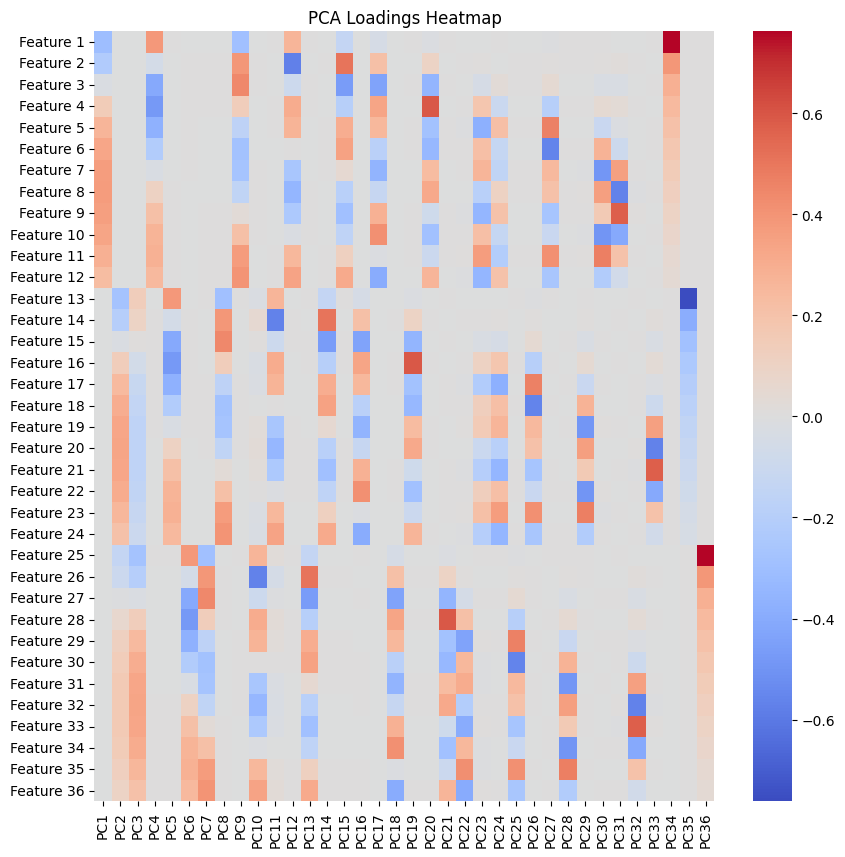}~~
\includegraphics[width=.3\textwidth]{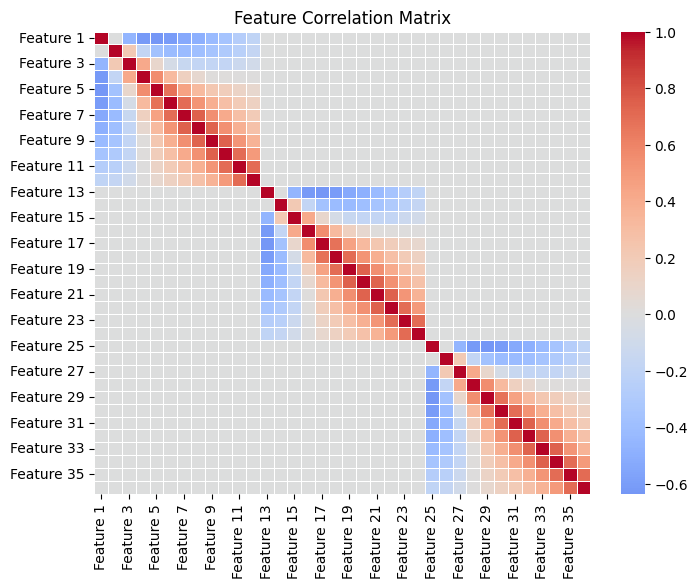}
\end{center}
\caption{\small\sf Explained variance ratio of Principal Components, a heatmap of their loadings, and the Spearman correlation matrix of the inputs for $n=12$.} \label{fig:Dataset} \end{figure}

The decision problem is uniquely determined by the triple of integer partitions $\mathbf t=(\lambda,\mu,\nu)$, which we represent as a $3n$-dimensional vector. PCA reveals a three-fold degeneracy in the principal components (PCs), arising from the symmetric role of $(\lambda,\mu,\nu)$ in the problem (recall Lemma \ref{lem-perm}), and a rapid drop in the explained variance ratio. Examining the Spearman correlation matrix also shows a negative correlation among the first few features and the rest, and a positive correlation among the last features. This stems from the lexicographic ordering and the constraint that the total sum of parts equals $n$ for each partition. We plot the explained variance of the PCs, their loadings, and the correlation matrix in Figure~\ref{fig:Dataset}.

\subsubsection{$18$-dimensional input} \label{subsub-18}
PCA, saliency analysis (see Section~\ref{sec:Saliency}), and the Kolmogorov--Arnold Network (see Section~\ref{sec:KAN}) all indicate that the first and last three entries of each partition vector are the most significant. Based on this observation, we conduct some of the experiments using only these 18 components instead of the full $3n$-dimensional vector.

\subsubsection{$1$-dimensional input} \label{subsub-1}
Finally, we conduct experiments using only the \( b \)-loading \( b(\mathbf{t}) \) for \( \mathbf{t} = (\lambda, \mu, \nu) \). We also tested using the triple \( (b_\lambda, b_\mu, b_\nu) \) instead of their sum \( b(\mathbf{t}) \). However, the accuracy remained unchanged, suggesting that the individual values \( (b_\lambda, b_\mu, b_\nu) \) do not carry more information than the sum.

\section{Saliency analysis of neural networks}
\label{sec:Saliency}

Even very simple neural networks (NNs) achieve relatively high accuracy on the binary classification task. We found that the results are not highly sensitive to architectural choices, such as the number of hidden layers, the number of nodes per layer, or the activation function. The plots below correspond to a NN with two hidden layers, each containing 16 nodes with \texttt{ReLU} activation, trained for 60 epochs with early stopping if the test set accuracy begins to diverge. The $3n$-dimensional dataset from \S \ref{subsub-3n} is used in this experiment. Training is performed using the Adam optimizer with cross-entropy loss. While the embedding dimension seems very small for this NN, we performed a PCA of the embedded features and found that the network effectively utilizes only a few dimensions. Depending on $n$, 80\% of the explained variance is captured within the first 3--7 PCs of the embedding vectors.

In order to understand what the algorithm bases its decision on, we perform a gradient saliency analysis, which means we compute the gradient of the NN output with respect to its input. Larger absolute values of the gradient indicate the final result depends more strongly on these particular input parameters. While we present the input as a $3n$-dimensional vector to the NN, we display the saliency map as a $3\times n$ matrix, where each row corresponds to one of the three integer partitions $(\lambda,\mu,\nu)$.  We also scale our input to zero mean and unit variance and use a train:test split of 90:10.

\subsection{Results}
\begin{figure}[t]
\centering
\includegraphics[width=\textwidth]{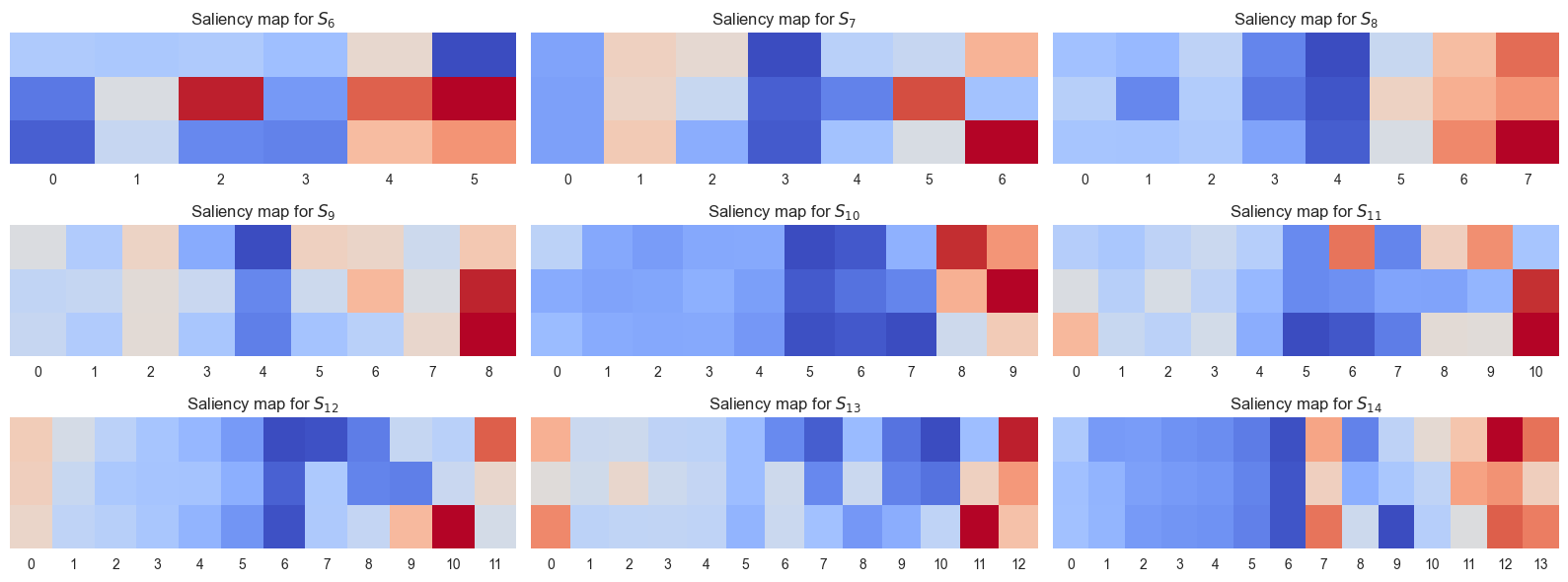}
\caption{\small\sf Saliency heatmap for the partition vectors from least relevant (cool blue) to most relevant (dark red).} \label{fig:Saliency} \end{figure}

The saliency analysis for $6\leq n \leq14$ is given in Figure~\ref{fig:Saliency}. Red indicates high saliency while blue indicates low saliency. We see that the middle entries of each partition are the least relevant for the decision. The asymmetry in the saliencies of the three partition vectors comes from the fact that the train and test set are selected randomly, which breaks their symmetric appearance.

Based on our analysis, we trained a classifier using only the first three and last three entries of each partition---corresponding to the dataset described in \S \ref{subsub-18}. This reduction had little to no impact on prediction accuracy, confirming the saliency analysis result that these entries are the most important.

Given that the dataset is somewhat unbalanced---the majority class comprises around 60\% of the dataset for larger $n$-values in the range we studied, though the majority class shifts between small and large $n$ (see the right plot of Figure~\ref{fig:Accuracy})---we also examined the confusion matrix. While the confusion matrix is skewed, it is not always towards the majority class. For example, for $n=8$, where the dataset is relatively balanced, 57\% of the misclassified Kronecker coefficients were misclassified as non-vanishing (but 47.5\% of all coefficients are non-vanishing), while for $n=14$, 53\% of the misclassified coefficients were misclassified as non-vanishing (and 67\% of all Kronecker coefficients are non-vanishing in this case).

\section{Kolmogorov--Arnold Networks (KAN)}
\label{sec:KAN}
KANs are by design more interpretable than neural networks and are well-suited for applications in scientific discovery~\cite{liu2024kan}. Overall, they are smaller models, and the non-linear functions on the edges of the computational graph, which we fit with cubic splines, allow for easy visualization. 

\subsection{Model architecture}
We find that the classification performance shows  architecture dependence. We investigate KAN architectures of shapes $[3n,\ell,1]$ and $[3n,\ell,\ell,1]$. We follow the notation of~\cite{liu2024kan}, which indicates that the KAN has a $3n$-dimensional input, one (respectively two) ``hidden layers'' of width $\ell$, and one scalar output. Throughout our experiments, we use cubic splines, a grid size of 5, and train for 200 steps with a batch size of 1024, binary cross entropy loss with logits (meaning a sigmoid activation function is applied to the output of the KAN), and standard parameters otherwise.

We also trained a $[1,1]$-KAN (essentially a cubic spline) to classify based on the dataset of $b$-loadings in \S \ref{subsub-1} and use KAN's \texttt{auto\_symbolic} function for symbolic regression.

\subsection{Results}
For the $[3n,\ell,1]$ KANs, we find that, after pruning, only a single node in the hidden layer is utilized. This aligns with our observation that the feed-forward NN also primarily utilizes a low-dimensional embedding manifold. Effectively, the KAN is learning a non-linear transformation of each entry in the partition vectors, which it then sums up and applies another non-linearity to obtain a prediction for the output. Accuracy percents are in the low 80's (for example, 83\% for $n=12$). 

For the $[3n,\ell,\ell,1]$ KANs, accuracy is higher (for example, 89\% for $n=12$). The KAN is utilizing more than one node in the first hidden layer, but still typically not all $\ell$ at its disposal. In the second hidden layer, it is utilizing even fewer. 

For both architectures, we see that the KAN is mostly paying attention to roughly the first half of the entries (in contrast to the NNs, where a saliency analysis showed that the NN is mostly paying attention to the first few and last few entries). The information contained in these entries is comparable, as knowing either the first half or the first and last few entries of a partition vector significantly constrains the remaining entries. Indeed, we observe that providing either set to the classifier results in accuracies in the mid 80\% range.

\begin{figure}[t]
\centering
\includegraphics[width=0.9\textwidth]{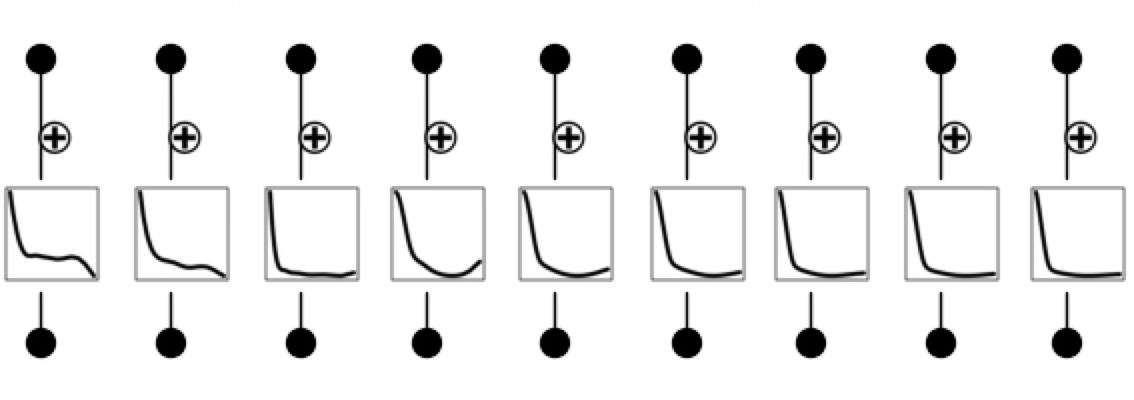}
\includegraphics[width=0.9\textwidth]{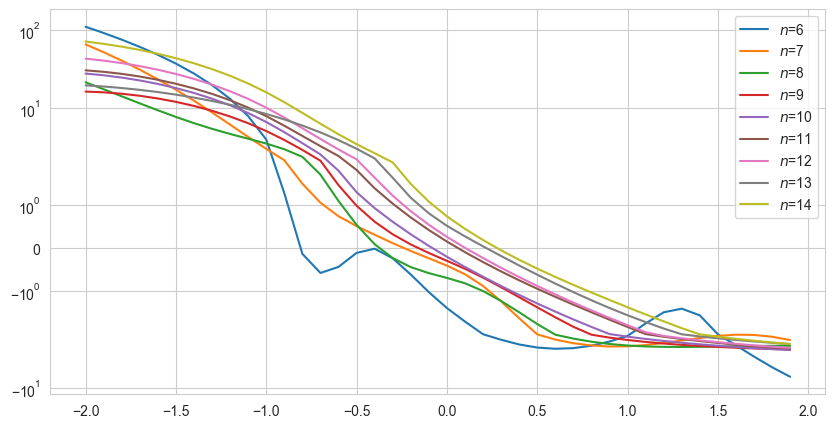}
\caption{\small\sf Top: [1,1] KANs for binary classification based on $b$-loadings for $6\leq n\leq14$. Bottom: Plot of the spline for $-2\leq x\leq2$.} 
\label{fig:KANFunctions}. 
\end{figure}

For the $[1,1]$-KAN, we find accuracies in the low to mid-80s, comparable to those in Figure~\ref{fig:Accuracy}. The function learned by the KAN to classify based on the $b$-loadings looks similar for all $n$; see Figure~\ref{fig:KANFunctions}. For binary classification with the sigmoid function
\begin{align}
    \sigma(x) \seteq \frac 1 {1+\exp(-x)} 
\end{align}
the decision boundary is at $x=0$ where $\sigma(0)=0.5$. In our $[1,1]$-KAN, we use $\sigma(f(x))$. While the optimizer will find an $f(x)$ that makes the transition between the classes sharper, the relevant property of $f$ for the final class assignment is where it changes sign. 

We train the KAN with normalized inputs $\tilde{b} \seteq \dfrac{b-\texttt{m}}{\texttt{s}}$, where $\texttt{m}$ is the mean of the distribution of the $b$-loadings and $\texttt{s}$ is the standard deviation. We find that all splines are monotonic for a wide range of $\tilde{b}$-values, and they have a unique zero close to $\tilde{b}=0$, i.e., close to $b=\texttt{m}$. Hence, when using $\tilde b$-values, we can approximate the spline by $\tilde f(x)\approx -x + c$ with a constant $c$. We observe from the spline that $c$ lies between $-0.5$ and $0.5$ and is increasing as $n$ increases. If we set $c=0$, this function predicts the zero/non-zero classes with accuracies in the low 80's range, which is comparable to those in Figure~\ref{fig:Accuracy}. Consequently,  
\begin{equation} \label{eq:KAN} \tilde F_1(\tilde b) = \sigma(-\tilde b) \qquad \text{ or } \qquad  F_1(b) = \sigma (-b+\texttt{m})  \end{equation}
serves as an effective decision function for $6 \le n \le 14$. Understanding why the mean $\texttt{m}$  of the $b$-loadings can act as a decision boundary would be an interesting direction for further investigation.

\section{Small Neural Networks}

\subsection{Model architecture}

The next interpretable ML model is a small neural networks (NNs) with one hidden layer containing \( k \) nodes, using \texttt{ReLU} activation for the hidden layer and a sigmoid activation for the output. Since there is no improvement in accuracy when using the triple \( (b_\lambda, b_\mu, b_\nu) \) as input instead of the sum \( b(\mathbf{t}) = b_\lambda + b_\mu + b_\nu \), $\mathbf t = (\lambda, \mu, \nu)$, we take \( b(\mathbf{t}) \) as the input, hence using the dataset in \S \ref{subsub-1}. Then the layers in the  model  are described by 
\[
\mathbb{R}  \rightarrow \mathbb{R}^k  \rightarrow \mathbb{R},
\]
with a total of \( 3k+1 \) parameters.  

To be more precise, let \( \sigma (x)  \) be the sigmoid function as before and \( r(x) \coloneqq \max (0, x) \)  the \texttt{ReLU} function. Then, the NN model is given by the function:  
\begin{equation} \label{eqn-snn}
F(x) = \sigma \left( \gamma_0 + \sum_{i=1}^k \gamma_i \ r(\alpha_i  + \beta_i x) \right),
\end{equation}
where \( \alpha_i, \beta_i, \gamma_i \) (\( i=1, 2, \dots, k \)) and $\gamma_0$ are the parameters (or weights).

\subsection{ML results}
We run the model in \eqref{eqn-snn} for $9 \leq n \leq 14$ with a 66:33 train-test split. Across all $n$ and various values of $k$, we obtain a test accuracy of approximately 83\%. For example, when $n=14$ and $k=7$, the model has 22 parameters and 
we can easily look at them; the matrix {\tiny $\begin{bmatrix}  \alpha_i \\ \beta_i \\ \gamma_i \end{bmatrix}_{1 \le i \le 7} $} along with $\gamma_0$ are given by

\tiny\begin{align*}   &  \begin{bmatrix} 0&0&0&-2.4159462& -2.8786569&2.3823507& -3.1325226 \\
 -1.1845468 & -2.146678  & -0.7272987 &  0.26937068&  0.32904944& 0.8263043 &  0.12642458
             \\-0.591498  & -0.35002795& -0.27688783& -0.52213764& -0.24775109&  0.27229485& -0.6243779
            \end{bmatrix}, \\ & \hskip 2.1 in \gamma_0=2.1719017.   
\end{align*}
\normalsize
Substituting these values into \eqref{eqn-snn} yields an accuracy of approximately 83\%. However, we observe that $r(\alpha_i+ \beta_i x) \equiv 0$ with $\alpha_i=0$ and $\beta_i<0$ for $i=1,2,3$, indicating that $k=4$ would have been sufficient.

Motivated by this observation, we examine the networks for different values of $k$ and find that logistic regression is sufficient. Indeed,  a logistic regression model achieves an accuracy of approximately 83\% across all $n$. For instance, when $n=14$, the decision function for a threshold of 0.5 is given by 
\begin{equation} \label{eq:SNN} F_2(b) = \sigma (-0.08550718 \times b +7.27970193), \end{equation}
with a decision boundary at approximately $b=85.14$ and an accuracy of about 84\%. In comparison, the decision boundary from \eqref{eq:KAN} is $b=\texttt{m} \approx 72.07$ with an accuracy 83\%. Examining all coefficients, we find that the best separation occurs near $b=80$, achieving  an accuracy of about 85\%.

\section{Symbolic Regression}

As another approach to interpretable ML for predicting whether $g(\mathbf t) \neq 0$ from the $b$-loading of the partition triple $\mathbf t=(\mu,\nu,\lambda)$, we perform symbolic regression on the neural networks (NNs). Specifically, we first train a fully-connected NN with $4$ hidden layers, each containing $128$ nodes. We alternate activation functions between $\texttt{tanh}$ and $\texttt{GeLU}$ across the hidden layers and apply the $\sigma$ activation function to the final $1$-dimensional output layer. The NN is trained over 40,000 epochs using an Adam optimizer with learning rate $10^{-3}$. Finally, we use the trained NN to generate data for symbolic regression, which we perform using PySR \cite{cranmer2023interpretable}.

\begin{figure}[htb]
    \centering
    \includegraphics[width=0.8\textwidth]{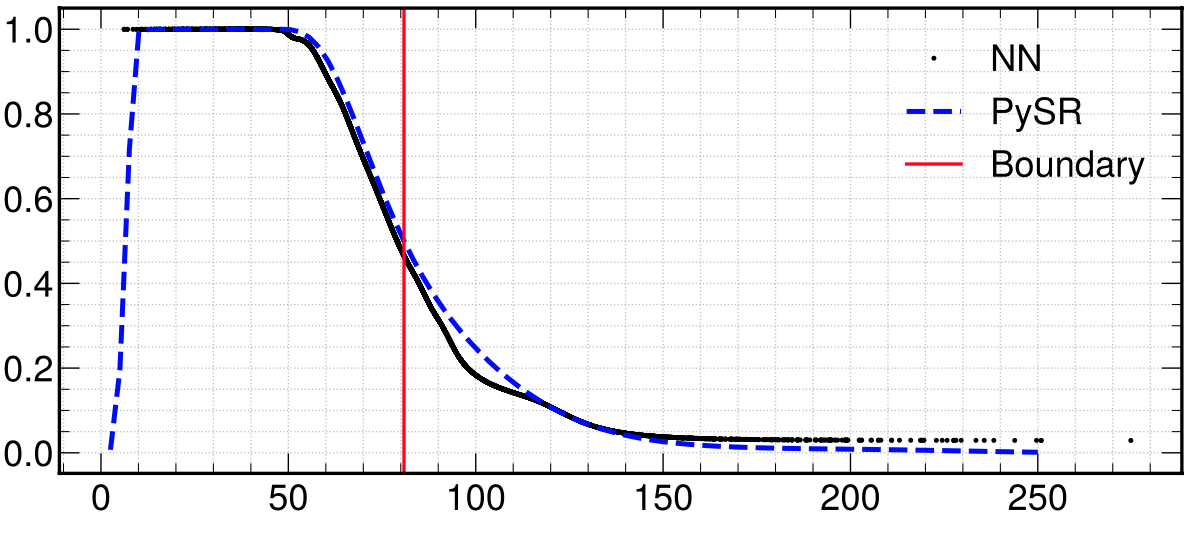}
    \caption{\small \sf Comparison of the expression \eqref{eq:symbolicBloading}, obtained through symbolic regression, with the output of a trained neural network.} \label{fig:sr}
\end{figure}

As a result, we identify the following candidate function, which approximates the NN model reasonably well:
\begin{gather}\label{eq:symbolicBloading}
    F_3(b) = \left[\cos\left(\sqrt{\cos(\sin(\log(b^2))^2)} + \cos(\log(b))\right)\right]^3.
\end{gather}
The predictor based on the condition $F_3(b)\geq 1/2$ achieves an accuracy of approximately 83\%. The performance is summarized in Figure~\ref{fig:sr}. In particular, the decision boundary is again around $b=80$.

\section{Transformers}
In this section, aside from the main theme of this article, we present ML results from transformer models, which achieve the highest reported accuracy of over 99\%. We switch back to the notation $g_{\mu,\nu}^\lambda= g(\mathbf t)$ for $\mathbf t= (\mu,\nu,\lambda)$. 

\medskip

\begin{figure}[t]
    \centering
    \begin{tikzpicture}
        \node at (0,0) {\includegraphics[width=1.0\textwidth]{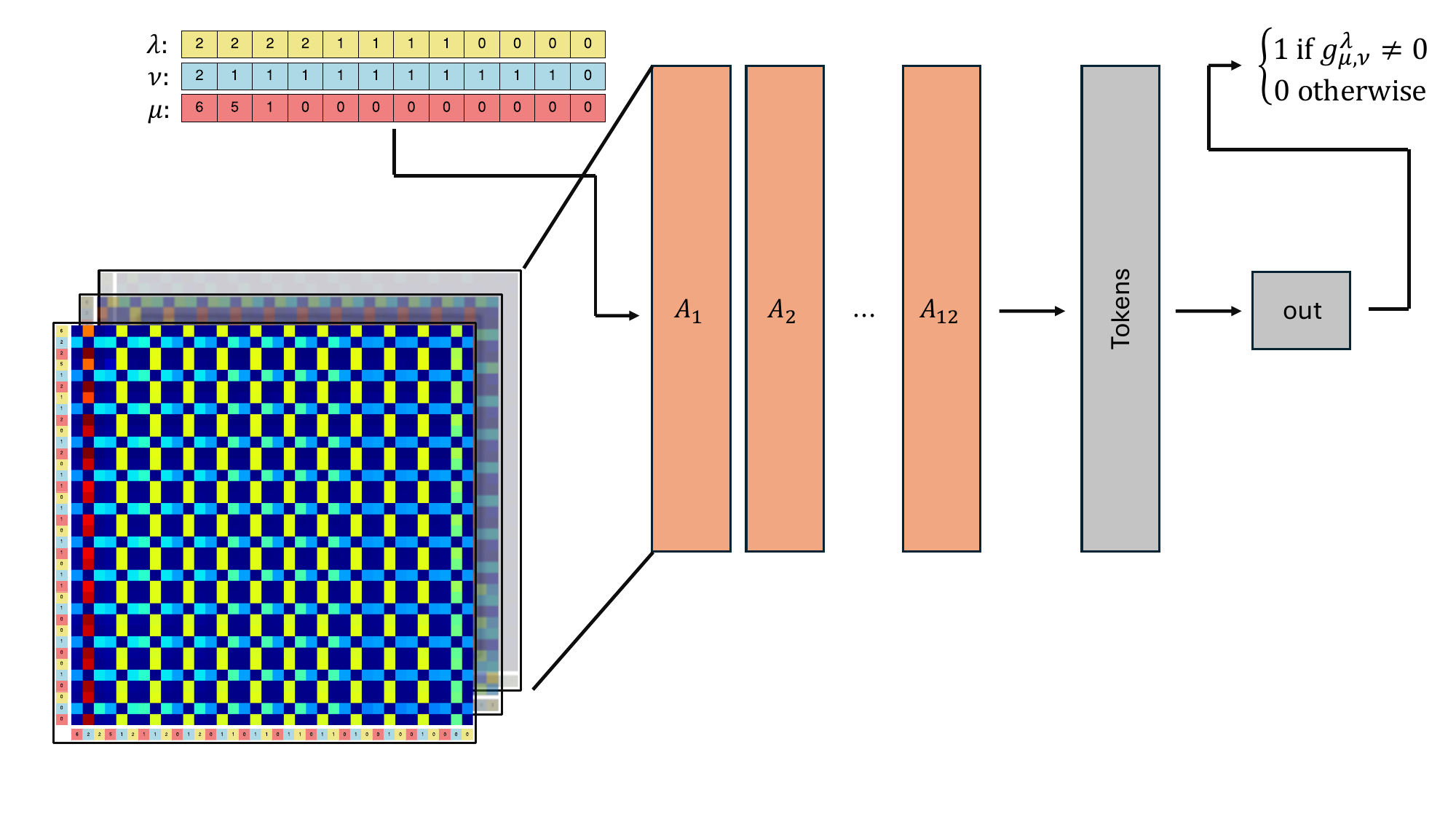}};
        \node at (-3.5, 1.5) {Attention matrices};
        \node at (0.75,-1.75) {$12$ Attention blocks};
    \end{tikzpicture}
    \caption{\small \sf Transformer architecture for deciding whether $g_{\mu,\nu}^\lambda \neq 0$ based on the partitions $(\mu,\nu,\lambda)$. The coloring of the input blocks $\mu,\nu$ and $\lambda$ corresponds to the positional encoding used in the transformer.}\label{fig:transformer}
\end{figure}

Evidently highly non-trivial combinations of the components of the partitions $\mu$, $\nu$ and $\lambda$ influence $g_{\mu,\nu}^\lambda$. Motivated with the success of the attention mechanism \cite{vaswani2017attention}, we use transformer-like architecture to build our model. In particular, we use $12$ multi-head attention blocks, each with $8$ heads. For a given $n$, we use $\{0,1,\dots, n\}$ as our token table, thus, each entry in a given partition $\mu$ is a token. Our context window is equal to $3n$. For the positional embedding, we use encoding that does not distinguish between individual parts of a given partition $\mu$. Due to this, we do not apply causal mask to the attention matrices. We do, however, distinguish different arguments $\mu,\nu$ and $\lambda$ of $g_{\mu,\nu}^\lambda$ by assigning them positions $0$, $1$ or $2$, respectively. The input to the transformer is then produced by interleaving position $0$, $1$ and $2$ of the stacked array $[ \mu,\nu, \lambda]$, as shown on Figure~\ref{fig:transformer}.

%GB: Not sure which version's better, including both for now. TODO remove old version.
% \begin{figure}[htb]
%     \centering
%     \begin{tikzpicture}
%         \node at (0,0) {\includegraphics[width=1.0\textwidth]{pic/transformer/transformer.pdf}};
%         \node at (0.8,2) {$12$ Attention blocks};
%     \end{tikzpicture}
%     \caption{Transformer architecture for predicting if $g_{\mu,\nu}^\lambda\neq 0$ from the partitions $(\mu,\nu,\lambda)$. The coloring of the blocks corresponds to the positional encoding used in the transformer.}\label{fig:transformer}
% \end{figure}

Given that the predicted object is $g_{\mu,\nu}^\lambda$, rather than a sequence, we modify the head of the transformer to apply over the full context window, thereby reducing the dimensionality of the output to $1$. The model is then trained as a binary classifier, with the classes being $g_{\mu,\nu}^\lambda = 0$ or $g_{\mu,\nu}^\lambda \neq 0$. We use binary cross-entropy as a loss function and train for $1.25\times 10^5$ epochs using an AdamW optimizer with no scheduler. We find that a learning rate of $\alpha = 3\times 10^{-4}$ with $\beta = (0.9, 0.95)$ and $\epsilon=10^{-8}$ performs reasonably well for the classification task. To ensure stability during training, we clip gradient norms to $1.0$. The training was performed on a single Nvidia P100 GPU, on which we can use a batch size of $128$. 

The accuracies on a test set (with a $33\%$ split) significantly outperform previous models \cite{L1}, as shown in Table~\ref{tab:transformerCompare}.

\begin{table}[t]
    \centering
    \begin{tabular}{c @{\hspace{10mm}} ccc}
    \toprule
    \multirow{2}{*}{$n$} & \multicolumn{3}{c}{Accuracy \%} \\
    \cmidrule(lr){2-4}
    & \small{Transformer} & \small{CNN} & \small{LGBM} \\
    \midrule
    12 & \bf{99.42} & 96.97 & 97.14\\
    13 & \bf{99.65} & 97.73 & 98.37\\
    14 & \bf{99.06} & 97.72 & 98.45\\
    \bottomrule
    \end{tabular}\vspace{0.3cm}
    \caption{\small \sf Performance of the transformer-like model for different values of $n$.}
    \label{tab:transformerCompare}
\end{table}

\section{Concluding Remarks}

We applied interpretable ML models to the decision problem of which Kronecker coefficients are vanishing. Using a $3n$-dimensional input, our saliency analysis and KANs indicate knowing either the first half or the first and last few entries of a partition vector is sufficient for classification. Indeed, an $18$-dimensional input that omits the middle part performed equally well. When using a one-dimensional input of $b$-loadings, our interpretable ML models revealed simple formulas, as shown in \eqref{eq:KAN}, \eqref{eq:SNN} and \eqref{eq:symbolicBloading}, all achieving approximately 83\% accuracy. 

These results also show the limitations of using a one-dimensional input such as the $b$-loadings. It would be ideal to find a simple decision function on two- or three-dimensional vectors with significantly higher accuracy. This suggests that developing additional informative features beyond the $b$-loading may be necessary.

Despite the decision problem being NP-hard, much better classifiers exist. We constructed transformer models and achieved the highest reported accuracy of over 99\%. Understanding the decision process of the transformer models---for example, through mechanistic interpretability---would be an interesting direction for future research.

Overall, the ability to analyze dataset saliency and derive simple decision functions highlights the importance and potential of interpretable ML in tackling research problems in mathematics.

\bibliographystyle{alpha}
\bibliography{references}
\newpage

\address{\hskip - 0.4 cm Department of Physics and Astronomy, University of New Hampshire, Durham, NH 03824, USA}

\noindent \email{\href{mailto:giorgi.butbaia@unh.edu}{giorgi.butbaia@unh.edu}}
\vskip 0.3 cm

\address{\hskip - 0.4 cm Department of Mathematics, University of Connecticut, Storrs, CT 06269, USA  \hfill \break Korea Institute for Advanced Study, Seoul 02455, Republic of Korea}

\noindent \email{\href{mailto:khlee@math.uconn.edu}{khlee@math.uconn.edu}}
\vskip 0.3 cm

\address{\hskip - 0.4 cm Department of Physics and Department of Mathematics,
Northeastern University, Boston, MA 02115, USA \hfill \break 
NSF Institute for Artificial Intelligence and Fundamental Interactions, Cambridge, MA 02139, USA}

\noindent \email{\href{mailto:f.ruehle@northeastern.edu}{f.ruehle@northeastern.edu}}

\end{document}